\documentclass[nohyperref]{article}

\usepackage{microtype}
\usepackage{graphicx}
\usepackage{subfigure}
\usepackage{booktabs} 

\usepackage{hyperref}

\usepackage[accepted]{icml2023}

\usepackage{amsmath}
\usepackage{amssymb}
\usepackage{mathtools}
\usepackage{amsthm}
\usepackage{dsfont}

\usepackage[capitalize,noabbrev]{cleveref}

\theoremstyle{plain}

\theoremstyle{definition}

\theoremstyle{remark}

\newcommand{\id}{\mathds{1}}

\renewcommand{\epsilon}{\varepsilon}
\newcommand{\eps}{{\varepsilon}}
\renewcommand{\phi}{\varphi}
\renewcommand{\theta}{\vartheta}

\newcommand{\R}{\mathbb{R}}

\newcommand{\N}{\mathbb{N}}

\newcommand{\ex}{\mathbb{E}}        

\newcommand{\Nc}{\mathcal{N}}

\newcommand{\Yc}{\mathcal{Y}}

\newcommand{\Uc}{\mathcal{U}}

\newcommand{\Xb}{\mathbf{X}}
\newcommand{\Yb}{\mathbf{Y}}

\DeclareMathOperator*{\argmin}{arg\,min}

\newcommand{\diff}{{\,\mathrm{d}}}       

\usepackage[textsize=tiny]{todonotes}

\icmltitlerunning{FNNs: Shift Invariant Models for Functional Data}

\begin{document}

\twocolumn[
\icmltitle{Functional Neural Networks: Shift Invariant Models for Functional Data with Applications to EEG Classification}



\icmlsetsymbol{equal}{*}

\begin{icmlauthorlist}
\icmlauthor{Florian Heinrichs}{comp}
\icmlauthor{Mavin Heim}{comp}
\icmlauthor{Corinna Weber}{comp}
\end{icmlauthorlist}

\icmlaffiliation{comp}{SNAP GmbH, Gesundheitscampus-Süd 17, 44801 Bochum, Germany}

\icmlcorrespondingauthor{Florian Heinrichs}{mail@florian-heinrichs.de, florian.heinrichs@snap-gmbh.com}
\icmlcorrespondingauthor{Mavin Heim}{mavin.heim@snap-gmbh.com}
\icmlcorrespondingauthor{Corinna Weber}{weber@snap-gmbh.com}

\icmlkeywords{Functional data, deep learning, neural networks, shift invariance, sliding windows}

\vskip 0.3in
]



\printAffiliationsAndNotice{} 

\begin{abstract}
It is desirable for statistical models to detect signals of interest independently of their position. If the data is generated by some smooth process, this additional structure should be taken into account. We introduce a new class of neural networks that are shift invariant and preserve smoothness of the data: functional neural networks (FNNs). For this, we use methods from functional data analysis (FDA) to extend multi-layer perceptrons and convolutional neural networks to functional data. We propose different model architectures, show that the models outperform a benchmark model from FDA in terms of accuracy and successfully use FNNs to classify electroencephalography (EEG) data.
\end{abstract}

\section{Introduction} \label{sec:intro}

When autonomous machines act in their environment or humans interact with computers, the necessary input data is often streamed continuously. In some settings, the input streams can be easily transformed into control signals based on simple physical models. However, in more advanced scenarios, it is necessary to develop a more complex, data-driven model and use its predictions to control the machine. An important example of the latter scenario are brain-computer interfaces (BCIs), which record the user's brain activity and decode control signals based on the measured data. 

Most statistical models require an input of fixed dimension and a common approach is to extract windows of a fixed size with a fixed step size from the continuous data stream. These \textit{sliding windows} are then used to predict the desired control signal with one classification per window. The advantage of this approach is that only the most recent data is used for predictions, but it comes at cost: the signal of interest might occur at any time point in the extracted window. Thus, any classifier of the sliding windows needs to be ``shift invariant'' in the sense that it detects the desired signal independently of its position in the window.

In the context of BCIs, more specifically for the analysis of data measured via electroencephalography (EEG), traditional methods are based on carefully selected features that are calculated from the data. Commonly applied techniques include the principal component analysis of EEG data in the time domain, or features based on the power spectrum in the frequency domain \citep{azlan2014, boubchir2017, boonyakitanont2020}. Due to recent advances in the field of deep learning, different architectures of neural networks were proposed that avoid a manual feature extraction and seem to outperform more traditional methods. For example the neural network \textit{EEGNet} was proposed to support multiple BCI paradigms and is often referred to as benchmark model in the field \cite{lawhern2018}. In a clinical setting, some variant of the VGG16 neural network was used to detect signals associated with epilepsy \cite{dasilva2021}. In general, deep learning has been applied successfully to a variety of tasks related to EEG data \citep{craik2019, roy2019}.

Inspired by their successes in computer vision and natural language processing, common neural networks used for the classification of EEG data are based on convolutions. Convolutional neural networks (CNNs) using some form of pooling can be shift-invariant and therefore a good choice in the given context. However, they do not take the specific structure of EEG data into account. Similar to most physical processes, the electrical activity, that is recorded on the user's scalp by the EEG, can be considered as smooth \citep{ramsay2005}.
This additional structure is not taken into account by traditional multivariate methods, including deep neural networks, and it might be more appropriate to model the data as a (discretized) sample of an underlying smooth function. With this approach new information becomes available. For example, the use of derivatives could uncover hidden patterns, while smoothing techniques can increase the signal-to-noise ratio. Further, it is generally easier to interpret smooth functions compared to high dimensional vectors.

The latter paradigm is the basis of \textit{functional data analysis} (FDA), a branch of statistics that received more and more attention throughout the last decades and remains an active area of research. Most concepts from multivariate statistics have been extended to functional data \citep{ramsay2005, kokoszka2017}. For example, many functional versions of principal component analysis have been proposed in literature \citep{shang2014}. Different generalizations of the linear model to functional covariates and/or functional responses have been introduced \citep{cardot1999, cuevas2002}. 
Finally, Portmanteau-type tests for detecting serial correlation have been proposed for functional time series \citep{gabrys2007, bucher2020}. This functional approach allows it to extract previously unavailable information from the data in form of derivatives of the continuous signal.
 
As methods from FDA take the functional structure of physical processes into account, they would be suitable classifiers. However, classic methods from FDA are in general not shift invariant and require the signal of interest to be at a fixed point in time. In some applications it is possible to \textit{register} the functions through a suitable transformation of time \citep{sakoe1978, kneip1992, gasser1995, ramsay1998}. However, in the context of sliding windows, curve registration is often not feasible and methodology that requires previous registration cannot be applied reliably.

In the present work, we propose a framework that combines the advantages of neural networks (particularly CNNs) and FDA: functional neural networks (FNNs). On one side, these networks are shift invariant, and on the other side, they are able to model the functional structure of their input. FNNs have several advantages over scalar-valued neural networks. They are independent of the sample frequency of the input data, as long as the input can be rescaled to a certain interval. Further, they allow to predict smooth outputs. And finally, they are more transparent to some extent due to smoothness constraints.

We summarize our contribution as follows:
\begin{itemize}
	\item We propose extensions of fully-connected and convolutional layers to functional data.
	\item We present architectures of functional neural networks based on these extensions.
	\item We show that the proposed methodology works through a simulation study and real data experiments.
\end{itemize}

Whereas multi-layer perceptrons (MLPs) are not shift invariant, the introduced functional convolutional layers allow the construction of shift invariant functional CNNs. This makes FNNs helpful in any scenario where sliding windows based on a (possibly multivariate) continuous data stream are classified, and they can be employed in a variety of applications.

\section{Related Work} \label{sec:rel_work}

To the best of our knowledge, the combination of functional data and (convolutional) neural networks is only discussed in a handful of papers and the proposed methodology extends previous results. In early works MLPs with functional inputs and neurons, that transform the functional data to scalar values in the first layer, were introduced \citep{Rossi2002, rossi2004, rossi2005}. Zhao proposed an algorithm to train similar MLPs with inputs from a real Hilbert space \yrcite{zhao2012}. Subsequently, Wang et~al. proposed to use functional principal components for the transformation of the functional inputs to scalar values in the first layer \yrcite{wang2019}. Wang et al. added another layer based on functional principal components to transform the scalar-valued output of the MLP back to functional data in the last layer \yrcite{wang2020}. More recently, fully functional neurons were proposed \citep{rao2021scalar, Rao2021}.

Besides these methods based on neural networks, a variety of approaches to classify functional data was proposed in the FDA literature. Most of these methods are extensions of their non-functional counterparts, including functional generalized linear models \citep{marx1999, cuevas2002, james2002, muller2005} and functional logistic regression \citep{wang2007, araki2009, rincon2012, berrendero2023}. Another approach to functional classification is to reduce the dimension of the data first, and subsequently use a classification method for multivariate data. Examples of this include FPCA, a variant of PCA for functional data \citep{hall2001, delaigle2012} and variable selection \cite{berrendero2016}.

Often the argument of the functional observations represents time. In this case, the data might be considered as time series instead of functions and a time series classifier might be used. Different architectures, including MLPs, CNNs and residual networks were proposed \citep{le2016, wang2017, serra2018}. Ismail Fawaz et~al. compared a variety of neural networks for time series classification \yrcite{ismail2019}. 

\section{Mathematical Preliminaries} \label{sec:math}
Let us assume, we observe $d$ quantities at $T$ time instants for $N\in\N$ individuals, providing us with matrices of observations 

$$ \Xb^{(n)} = \left(\begin{array}{ccc}
	X_{1,1}^{(n)} & \cdots & X_{1,T}^{(n)} \\
	\vdots & \ddots & \vdots \\
	X_{d, 1}^{(n)} & \cdots & X_{d,T}^{(n)}
\end{array}\right),
$$
for $n=1,\dots, N$, and jointly with $\Xb^{(n)}$ their corresponding \textit{labels} $\Yb^{(n)}$ which might be vectors in $\R^c$ or matrices 
$$ \Yb^{(n)} = \left(\begin{array}{ccc}
	Y_{1,1}^{(n)} & \cdots & Y_{1,T}^{(n)} \\
	\vdots & \ddots & \vdots \\
	Y_{c, 1}^{(n)} & \cdots & Y_{c,T}^{(n)},
\end{array}\right)
$$
where $c$ denotes the number of quantities that we observe for $\Yb^{(n)}$. Further assume the observed quantities to be noisy versions of an underlying smooth signal, i.\,e.,
\begin{equation}\label{loc-scale-model}
	X_{i,t}^{(n)}=f_i^{(n)}\big(\tfrac tT \big) + \eps_{i,t}^{(n)},
\end{equation} 
for smooth functions $f_i^{(n)}$ and centered errors $\eps_{i,t}^{(n)}$, for $i=1,\dots, d, t=1,\dots, T$ and $n=1,\dots, N$. Note that the degree of smoothness might vary for different applications, which leads to slight modifications in the model. This representation suggests the use of methods from functional data analysis, which consider the intrinsic structure of the data. 

Throughout this work, we only require the functions $f_i^{(n)}$ to be square-integrable, i.\,e., $f_i^{(n)} \in L^2([0, 1])= \{ f:[0, 1] \to \R | \int_0^1 f^2(x) \diff x < \infty \}$.  
Similarly, in case of matrix-valued labels $\Yb^{(n)}$, we assume their entries to be discretized versions of some underlying functions $g_i^{(n)} \in L^2([0,1])$, more specifically $Y_{i,t}^{(n)} = g_i^{(n)}(t/T)$. 

Our aim is to approximate the functional $F:\big(L^2([0,1])\big)^d \to \Yc$, which maps an observation $\Xb$ to its corresponding label $\Yb$, where $\Yc = \R^c$ for vector-valued labels $\Yb^{(n)}$ and $\Yc = \big(L^2([0,1])\big)^c$ for matrix-valued labels. Formally, the functional $F$ corresponds to the conditional expectation $\ex[\Yb|\Xb]$.

In case of classification problems, each coordinate $F_i(\Xb)$ of the functional $F$ can be interpreted as the probability of $\Xb$ belonging to class $i \in \{1,\dots, c\}$.

\subsection{Preprocessing}
Before putting observations into a neural network, it is often helpful to preprocess them by applying certain filters or normalization. In our case, we work with noisy functional data, observed at discrete time points. In functional data analysis, a common first step for this kind of data is \textit{smoothing}, which helps to reduce the errors and extends the observations from discrete time points to a continuous interval. Another preprocessing step frequently used for neural networks, is some form of normalization to ensure that the data is of a similar magnitude. We employ \textit{local linear estimation} for smoothing the data and \textit{standardization} for its normalization as described below.

\subsubsection{Smoothing}\label{sec:preprocessing}
In the literature, there exists a variety of smoothing procedures from Fourier series to expansions based on B-splines or wavelets \citep{ramsay2005}. We use local polynomial regression to estimate the functions $f_i^{(n)}$ and their first derivative(s) \citep{fan1996}. 

For the sake of clarity, we omit some indices and rewrite \eqref{loc-scale-model} as $X_{t}=f\big(\tfrac tT \big) + \eps_{t}$ for a moment. Then, if $f$ is $p+1$ times differentiable with bounded derivatives, we can define the local polynomial estimator as 
\begin{align}\label{eq:loc_lin}
&(\hat{f}(x), \widehat{f'}(x), \dots, \widehat{f^{(p)}}(x))\\ \nonumber &= \argmin_{\beta_0, \dots, \beta_p} \sum_{t=1}^{T} \Big(X_t - \sum_{j=0}^{p} \beta_j \big(\tfrac{t}{T}-x\big)^j\Big)^2 K_h\big(\tfrac{t}{T}-x\big)
\end{align}
to estimate $f$ and its first $p$ derivatives. Here $K$ denotes a kernel function, $h$ the bandwidth of the estimator and $K_h(\cdot)=K(\tfrac{\cdot}{h})$. In the following, we assume $K:\R\to\R$ to be a symmetric, twice differentiable function, supported on the interval $[-1, 1]$ and satisfying $1 = \int_{-1}^1K(x)\diff x > \int_{-1}^1x^2 K(x)\diff x$ . 

From the above definition, explicit formulas can be derived for the estimators by setting the derivatives of the right-hand side with respect to $\beta_j~(j=1,\dots, p)$ to zero and solving the resulting system of linear equations. Exemplary calculations for the local linear case, where $p=1$, are provided in Appendix \ref{app:loc_lin}.

To simplify the notation, we will refer to the estimators of the functions $f_i^{(n)}$ and their derivatives as $h_{i, 1}^{(n)}$, thus, we obtain estimators $(h_{i, 1}^{(n)}, (h_{i, 1}^{(n)})', \dots, (h_{i, 1}^{(n)})^{(p)})$ for each $f_i^{(n)}$, $i=1,\dots, d, n=1,\dots, N$.

The choice of the bandwidth is crucial in order to obtain a good estimate of the underlying functions. If the bandwidth is chosen too small, the estimator will overfit the data, whereas a large bandwidth leads to over-smoothing \citep{silverman2018}. Oftentimes it is a good idea to use cross validation to select a bandwidth that minimizes a certain error measure, such as the mean squared error. Generally, the estimation of higher derivatives requires larger bandwidths than the estimation of the function itself.

\subsubsection{Normalization}
When neural networks are trained via some form of gradient descent, it is crucial to ensure that the input data is of a similar size, which is done through prior normalization. There are many different normalization methods and the most useful choice depends on the specific application. In the following, we will standardize the data by subtracting the mean and dividing by the standard deviation across a suitable range of the data. As we did not make any assumptions about the relation between the signals $f_i^{(n)}$ and $f_j^{(n)}$, we standardize each smoothed signal $h_{i, 1}^{(n)}$ (and its derivatives) separately, i.\,e., we calculate
$$h_{i, 2}^{(n)} = \frac{h_{i, 1}^{(n)} - \int_0^1 h_{i, 1}^{(n)}(x)\diff x}{\Big(\int_0^1 \big(h_{i, 1}^{(n)}(x) - \int_0^1 h_{i, 1}^{(n)}(y)\diff y\big)^2 \diff x\Big)^{1/2}}. $$
After this transformation, the signals are of a similar magnitude, for each observation $\Xb^{(n)}$.

\section{Functional Layers}  \label{sec:dev1}

\subsection{Functional Multilayer Perceptrons}
\label{sec:f_mlp}
Once the data is smoothed and prepared to be analyzed as functional data, it is not clear how to design neural networks that take this additional structure into account. The simplest form of an artificial neural network with scalar input $(h_1, \dots, h_d)$ is the multilayer perceptron (MLP), that consists of $L$ layers with $J_1, J_2, \dots, J_L$ neurons each. The value at neuron $k$ in the $\ell$-th layer is then calculated as 
\begin{equation*}
	H_{k}^{\ell} = \sigma\bigg( b_{k}^{\ell} + \sum_{j=1}^{J_{\ell-1}}  w_{j,k}^{\ell}  H_{j}^{\ell-1}\bigg),
\end{equation*}
where $H_{k}^{0}=h_k$ denotes the network's input, $H_{k}^{L}$ its output, $b_{k}^{\ell}$ the $k$th neuron's bias and $w_{j,k}^{\ell}$ the weight between the $j$th neuron in layer $\ell-1$ and the $k$th neuron in layer $\ell$. The function $\sigma:\R\to\R$ is referred to as \textit{activation function} and enables the network to reflect non-linear dependencies.

When the input data is not scalar, but functional, Rao and Reimherr \yrcite{Rao2021} propose to replace the scalar biases by functional biases and the weights between neurons by integral kernels, finally defining the neurons' values as
\begin{equation}\label{eq:neurons_rao}
	H_{k}^{\ell}(s) = \sigma\bigg( b_{k}^{\ell}(s) + \sum_{j=1}^{J_{\ell-1}} \int  w_{j,k}^{\ell}(s, t)  H_{j}^{\ell-1}(t) \diff t \bigg).
\end{equation} 
While the use of integral kernels $w_{j, k}^{\ell}\in L^2([0,1]^2)$ allows to model rather general relations between the model's input and its desired output, this flexibility makes the network's training difficult because we need to find optimal weight functions for any connection between two neurons. 

Starting from the fully-connected MLP, many advances in deep learning are due to more specific architectures, which reduce the number of model parameters to mitigate the curse of dimensionality. For instance, CNNs can be interpreted as MLPs, where most weights vanish and the remaining connections between neurons share a smaller set of weights. 

In a similar fashion, we propose to simplify the neuron model in \eqref{eq:neurons_rao} by using weight functions $w_{j,k}^{\ell}:[0, 1]\to\R$ rather than integral kernels in $L^2([0,1]^2)$. This adaptation leads to neurons defined via 
\begin{equation}\label{eq:neurons}
	H_{k}^{\ell}(t) = \sigma\bigg( b_{k}^{\ell}(t) + \sum_{j=1}^{J_{\ell-1}} w_{j,k}^{\ell}(t)  H_{j}^{\ell-1}(t) \bigg).
\end{equation}
The above defined neurons are fully functional in the sense that both their input and output are functions. If we try to predict scalar-valued labels in $\R^c$, we need to summarize the information contained in the functions. We propose to calculate the scalar product of the weights and their corresponding inputs, leading to
\begin{equation}\label{eq:neurons_last_layer}
	H_{k}^{\ell} = \sigma\bigg( b_{k}^{\ell} + \sum_{j=1}^{J_{\ell-1}} \int w_{j,k}^{\ell}(t)  H_{j}^{\ell-1}(t) \diff t \bigg).
\end{equation}
The scalar product of functions $f, g \in L^2([0,1])$ is simply the integral $\int_0^1 f(x) g(x)\diff x$ and plays the same role as average pooling in the case of conventional neural networks. With this definition of a functional multilayer perceptron (F-MLP), we simplified the training and need to optimize functional weights of a single variable. The theoretical framework to train the model through backpropagation based on Fréchet derivatives is provided by \citep{Rossi2002, Olver2016, Rao2021}. 

The computation of Fréchet derivatives becomes tedious and computationally expensive. An efficient approach to simplify computations and simultaneously reduce the dimension of the weights' space, is to replace the weights $w_{j,k}^{\ell}(t)$ by linear combinations of a finite set of base functions. Therefore, let $\{\phi_i\}_{i=0}^q$ be a set of suitable functions, such as Legendre polynomials, wavelets or the first $q/2$ sine-cosine pairs of the Fourier basis, and consider the linear combination
\begin{equation}\label{eq:base_rep}
	w_{j,k}^{\ell}(t) = \sum_{i=0}^{q} w_{j,k}^{\ell, i} \phi_i(t),
\end{equation} 
for some scalar weights $w_{j,k}^{\ell, i}$. With this representation, the fully functional neural network can be described through scalar weights and we are able to use the standard scalar backpropagation.

\subsection{Functional Convolutional Neural Networks}

The F-MLP is particularly useful if the input functions are aligned (or can be aligned via a suitable transformation of time) and the signals of interest happen at the same time instants. However, under the sliding window paradigm, for high-noise data such as speech or EEG signals, it is not possible (or at least not useful) to previously register the curves, as the signal of interest may occur at any arbitrary time. In this case, MLPs are impractical as they would require many parameters to model complex patterns.

For scalar input, alternative architectures have been developed that are shift invariant and therefore capable to detect certain signals independently of their position. One type of neural network that is considered as ``translation invariant'' are CNNs, which we can extend to functional data as well.

Similarly to \eqref{eq:neurons_rao}, we can define a functional convolutional layer by setting $w_{j, k}^{\ell}(s, t) = u_{j, k}^{\ell}(s - t)$ for some filter (or kernel) function $u_{j, k}^{\ell}:\R\to\R$ with support on $[-b, b]$ and bandwidth $b \in (0, 1)$, ultimately leading to 
\begin{equation*}
	H_{k}^{\ell}(s) = \sigma\bigg( b_{k}^{\ell}(s) + \sum_{j=1}^{J_{\ell-1}} \int  u_{j, k}^{\ell}(s - t)  H_{j}^{\ell-1}(t) \diff t \bigg),
\end{equation*}
where the functions $H_{k}^{\ell}$ are extended to $[-b, 1 + b]$ by defining them as zero outside of the interval $[0, 1]$.

These functional convolutional layers are shift invariant in the sense that a filter, which is capable of detecting a certain signal, would detect it independently of its position in the interval $[0, 1]$. Once again, we reduce the dimension of the optimization problem by representing the filters as linear combinations of a set of base functions as in \eqref{eq:base_rep}.

\subsection{Architecture}\label{sec:architectures}

With functional versions of fully connected and convolutional layers at hand, we can define arbitrary architectures of functional neural networks (FNNs). Figure \ref{fig:architecture} displays FNNs with scalar and functional outputs, respectively. In both cases, the first layer uses local linear estimation to smooth the input and estimate derivatives of the smoothed signals, while the second layer standardizes the input across each signal. Following are two functional convolutional layers. For the FNN with scalar output, the last layer is a functional fully connected layer, while for the FNN with functional output, the last layer is a third functional convolutional layer.

\begin{figure}
	\vskip 0.1in
	\begin{center}
		\centerline{\frame{
\begin{tikzpicture}
		
	\node [draw,
	fill=black!5!white,
	minimum width=2.5cm,
	minimum height=0.6cm,
	rounded corners=2,
	]  (llelayer) at (2,0){LLE Layer};
	
	\node [draw,
	fill=black!5!white,
	minimum width=2.5cm,
	minimum height=0.6cm,
	rounded corners=2,
	below=0.35cm of llelayer
	]  (normalizer) {Standardization};
	
	\node [draw,
	fill=black!15!white, 
	minimum width=2.5cm,
	minimum height=0.6cm,
	rounded corners=2,
	below=0.35cm of normalizer
	] (conv1) {FuncConv};
	
	\node [draw,
	fill=black!15!white, 
	minimum width=2.5cm,
	minimum height=0.6cm,
	rounded corners=2,
	below=0.35cm of conv1
	] (conv2) {FuncConv};
		
	\node [draw,
	fill=black!25!white, 
	minimum width=2.5cm,
	minimum height=0.6cm,
	rounded corners=2,
	below=0.35cm of conv2
	] (dense) {FuncDense};
	
	\draw[-latex] (llelayer.south) -- (normalizer.north);
	\draw[-latex] (normalizer.south) -- (conv1.north);
	\draw[-latex] (conv1.south) -- (conv2.north);
	\draw[-latex] (conv2.south) -- (dense.north);
	
	\draw[latex-] (llelayer.north) -- (2,0.5)
	node[above]{Functional Input};
	\draw[-latex] (dense.south) -- +(0,-0.3) 
	node[below]{Scalar Output};
	
	\draw (4,1.2) -- (4,-5);
	\path (0,0) -- (8,0);
	
	\node [draw,
	fill=black!5!white,
	minimum width=2.5cm,
	minimum height=0.6cm,
	rounded corners=2,
	]  (llelayer) at (6,0){LLE Layer};
	
	\node [draw,
	fill=black!5!white,
	minimum width=2.5cm,
	minimum height=0.6cm,
	rounded corners=2,
	below=0.35cm of llelayer
	]  (normalizer) {Standardization};
	
	\node [draw,
	fill=black!15!white, 
	minimum width=2.5cm,
	minimum height=0.6cm,
	rounded corners=2,
	below=0.35cm of normalizer
	] (conv1) {FuncConv};
	
	\node [draw,
	fill=black!15!white, 
	minimum width=2.5cm,
	minimum height=0.6cm,
	rounded corners=2,
	below=0.35cm of conv1
	] (conv2) {FuncConv};
	
	\node [draw,
	fill=black!15!white, 
	minimum width=2.5cm,
	minimum height=0.6cm,
	rounded corners=2,
	below=0.35cm of conv2
	] (conv3) {FuncConv};
	
	\draw[-latex] (llelayer.south) -- (normalizer.north);
	\draw[-latex] (normalizer.south) -- (conv1.north);
	\draw[-latex] (conv1.south) -- (conv2.north);
	\draw[-latex] (conv2.south) -- (conv3.north);
	 
	\draw[latex-] (llelayer.north) -- (6,0.5)
	node[above]{Functional Input};
	\draw[-latex] (conv3.south) -- +(0,-0.3) 
	node[below]{Functional Output};
	
\end{tikzpicture}
}}
		\caption{Left: Neural network architecture with functional input and scalar output. Right: Neural network architecture with functional input and output.}
		\label{fig:architecture}
	\end{center}
	\vskip -0.2in
\end{figure}

\section{Empirical Results}  \label{sec:emp_res}

We now show that the proposed methodology works with simulated and real data and compare it to benchmark models.

\subsection{Simulation Study}
\label{sec:sim_study}
\subsubsection*{Setup} Inspired by brain activity measured through electroencephalography (EEG), we generate two data
sets for the simulation study. In both cases, we simulate two-dimensional samples that belong to one of three classes.

For dataset I, we draw independently frequencies $\alpha_n\sim \Uc_{[8, 12]}, \beta_n\sim \Uc_{[13, 30]}$, time shifts $t_1^{(n)}, t_2^{(n)} \sim \Uc_{[0,1]}$, class labels $c_n\sim\Uc_{\{1, 2, 3\}}$ and errors $\eps_{i,t}^{(n)}\sim\Nc(0, 1)$. Based on these random quantities, we construct the continuous signals
\begin{align*}
	f_i^{(n)}(x) = &(1-\gamma_i(c_n)) \cdot \sin(2\pi \alpha_n (x + t_i^{(n)})) \\&+ \gamma_i(c_n) \cdot \sin(2\pi \beta_n (x + t_i^{(n)}))
\end{align*}
with class dependent coefficients $\gamma(1) = (0, 0), \gamma(2) = (0.8, 0.4)$ and $\gamma(3) = (0.4, 0.8)$
and finally define the discretized, noisy samples $X_{i, t}^{(n)} = f_i^{(n)}(t/T) + \eps_{i,t}^{(n)} $, for $t=1, \dots, T$. Examples of each class are displayed in Figure \ref{fig:dataset1} of Appendix \ref{app:sim_data}.

For dataset II, we draw independently scaling factors $w_n\sim \Uc_{[0.05, 0.1]}$, time points  $t_n\sim \Uc_{[0,1]}$, class labels $c_n\sim\Uc_{\{1, 2, 3\}}$ and standard normally distributed errors $\eps_{i,t}^{(n)}\sim\Nc(0, 1)$. Based on the scaling factors $w_n$ and time points $t_n$, we construct continuous signals $$f^{(n)}(x) = \max\big\{ - \tfrac{4}{w_n^2} (x-t_n)^2 + 3 , 0 \big\},$$
which resemble spikes, as displayed in Figure \ref{fig:dataset2} of Appendix \ref{app:sim_data}. Again, we define discretized, noisy samples $X_{i, t}^{(n)} = \gamma_i(c) \cdot f^{(n)}(t/T) + \eps_{i,t}^{(n)}$, where the class dependent coefficients are $\gamma(1) = (0, 0), \gamma(2) = (1, 0)$ and $\gamma(3) = (0, 1)$.

For both datasets, we vary the sample size $N \in \{1000, 2000, 3000, 4000, 5000\}$, while keeping $T=250$ fixed. As benchmark models, we use (i) $k$-nearest neighbors (kNN) with a varying number of neighbors $k \in \{1, 2, ..., 19\}$, (ii) an MLP with 3 hidden layers of 10, 20 and 40 neurons and ReLU activation, and (iii) a CNN with 3 convolutional layers of 10, 20 and 20 filters and filter length 15, max pooling layers after the first two convolutional layers and pooling size 3, and finally two dense layers of 40 and 3 neurons with ReLU and softmax activation. Before feeding the data into the kNN model, we smooth it by applying a local linear estimator as described in Section \ref{sec:preprocessing}. 

To show that functional neural networks work, and indeed surpass the performance of the benchmark models, we use an FNN as described in Section \ref{sec:architectures} with two functional convolutional and one functional dense layer. For the local linear estimation, we use the quartic kernel $K(x)=\tfrac{15}{16}(1-x^2)^2$ with support $[-1, 1]$ and the bandwidth $h=5$ for the estimation of the smooth function and $h=10$ for the estimation of its derivative. For each functional layer, we used the first 5 Legendre polynomials as base functions, i.e. $\varphi_0(x)=1, \varphi_1(x)=x, \varphi_2(x) = \tfrac{1}{2}(3x^2-1), \varphi_3(x)=\tfrac{1}{2}(5x^3-3x)$ and $\varphi_4(x)=\tfrac{1}{8}(35x^4-30x^2+3)$. Further, we used 20 and 10 filters of size 25 for the two convolutional layers. As activation function we chose the \textit{exponential linear unit} (ELU), which is defined as $\sigma(x) = x \cdot \id(x\ge 0) + (\exp(x) - 1) \cdot \id(x<0)$. As loss function, we used the categorical crossentropy.

We trained each model 100 times, while generating a new dataset for each trial. The neural networks were trained with 5 epochs. We used the Adam optimizer with its default hyperparameters in TensorFlow, i.\,e., a learning rate of 0.001, $\beta_1=0.9, \beta_2=0.999$ and $\eps=10^{-7}$. Further, we used a standard normal distribution to initialize the weights. However, preliminary experiments with other distributions (uniform on $[-0.05, 0.05]$, He normal and Glorot uniform) suggested that the training is robust with respect to different choices of the initialization.

\subsubsection*{Results}
The results of the kNN model for both datasets with a varying number of samples $N$ and neighbors are displayed in Figure \ref{fig:simulation_results} of Appendix \ref{app:sim_data_res}. In Table \ref{tab:simulation_results}, the results of the kNN model with the best choice of neighbors $k$ is compared to the results of the neural networks. In all cases, the classifications of the FNN are more reliable than those of the benchmark models. The FNN achieved an accuracy above 99.6\% in all cases, whereas the kNN classifier achieved between 93.0\% and 99.3\% for dataset I and between 76.9\% and 86.4\% for dataset II. The CNN performed only slightly worse than the FNN with accuracies ranging from 93.5\% to 98.8\% and from 91.3\% to 99.2\% for datasets I and II respectively. The MLP did not yield competitive results for dataset I (37.9\%-57.9\%) and still lagged behind for dataset II (79.0\%-90.7\%). As expected, the shift invariance of the FNN makes it particularly helpful for dataset II, where the signal of interest may occur at any point in the observed interval.
\begin{table}[t]
	\caption{Mean accuracy of the classifiers for the simulated datasets.}
	\label{tab:simulation_results}
	\vskip 0.1in
	\begin{center}
		\begin{small}
			\begin{tabular}{ c | c c c | c }
				\toprule
				$N$ & KNN & CNN & MLP & FNN \\
				\midrule
				\multicolumn{5}{l}{\textit{Dataset} {\sc I}}\\
				1000 & 93.0\% & 93.5\% & 37.9\% & {99.6\%} \\
				2000 & 97.3\% & 96.6\% & 42.9\% & {99.8\%} \\
				3000 & 98.5\% & 98.1\% & 47.8\% & {99.8\%} \\
				4000 & 99.1\% & 98.1\% & 52.3\% & {99.7\%} \\
				5000 & 99.3\% & 98.8\% & 57.9\%  & {99.8\%}\\
				\midrule
				\multicolumn{5}{l}{\textit{Dataset} {\sc II}}\\
				1000 & 76.9\% & 91.3\% & 79.0\%  &  {99.6\%}\\
				2000 & 81.4\% & 97.0\% & 86.8\%  &  {99.8\%}\\
				3000 & 83.6\% & 98.4\% & 88.9\%  &  {99.8\%}\\
				4000 & 85.5\% & 99.0\% & 90.1\%  &  {99.9\%}\\
				5000 & 86.4\% & 99.2\% & 90.7\%  &  {99.9\%}\\
				\bottomrule
			\end{tabular}
		\end{small}
	\end{center}
	\vskip -0.2in
\end{table}

\subsection{Real Data Experiments (Functional Data)}
The first two datasets with real data are classically used to benchmark new methods in the field of FDA. The \textit{phoneme} dataset contains 2000 pairs of (discretized log-) periodograms and class membership to one of five phonemes and is extracted from the TIMIT database \citep{hastie1995}. The \textit{tecator} dataset contains 215 spectrometric curves of meat samples based on near infrared absorbance with the aim of predicting its fat content \citep{thodberg2015}. In principle, the tecator dataset defines a regression problem and we predicted the fat content (see MSE results in Table \ref{tab:tecator}). In the literature, however, the data is often split into two categories: those meat samples with fat content $<$20\% and $\ge$ 20\%, and we used these categories to define a classification problem. We used 80\% of each dataset for training and the remaining data for validation. The results of the validation data for the phoneme and the tecator dataset are displayed in Tables \ref{tab:phoneme} and \ref{tab:tecator}, respectively. 

For the phoneme dataset, the two best models achieve an accuracy of 91.53\% and outperform the CNN and MLP benchmarks. With 87.83\%, the EEGNet, a CNN designed for the analysis of EEG data, performs reasonably well on the dataset although it clearly was not designed for this type of data. Recently Berrendero et~al. combined various variable selection methods with different classifiers in an extensive simulation study and reported best classification accuracies of 99.53\% and 81.14\% for the tecator and phoneme dataset respectively \yrcite{berrendero2016}. For both datasets, FNNs outperform the compared FDA methods.
\begin{table}
	\caption{Mean accuracy of the classifiers for the phoneme dataset.}
	\label{tab:phoneme}
	\vskip 0.1in
	\begin{center}
		\begin{small}
			\begin{tabular}{ c | c c c }
				\toprule
				Model & Accuracy & Recall & Precision \\
				\midrule
				EEGNet & 87.83 & 88.01 & 88.22 \\
				CNN & 83.73 & 83.83 & 85.01 \\
				MLP & 79.95 & 80.55 & 80.05 \\
				FNN(40, 20) & 90.26 & 89.95 & 89.90 \\
				FNN(5, 10) & 89.12 & 89.45 & 89.52 \\
				FNN(3, 12) & 91.03 & 90.45 & 90.72 \\ 
				FNN(10) & \textbf{91.53} & 91.42 & 91.56 \\
				FNN(20) & \textbf{91.53} & \textbf{91.52} & \textbf{91.72} \\
				\bottomrule
			\end{tabular}
		\end{small}
	\end{center}
	\vskip -0.1in
\end{table}
\begin{table}
	\caption{Mean accuracy of the classification / regression models for the tecator dataset.}
	\label{tab:tecator}
	\vskip 0.1in
	\begin{center}
		\begin{small}
			\begin{tabular}{ l | c c c | c }
				\toprule			
				Model & Accuracy & Recall & Precision & MSE \\
				\midrule
				CNN & 73.56 & 71.77 & 81.32 & 132.72 \\
				MLP & 85.30 & 83.80 & 82.31 & 147.07 \\
				FNN(40, 20) & 98.96 & 98.39 & 99.24 & \textbf{1.86} \\
				FNN(5, 10) & 97.91 & 96.52 & 98.55 & 2.17 \\
				FNN(3, 12) & 97.65 & 97.02 & 98.13 & 2.93 \\
				FNN(10) & \textbf{100.00} & \textbf{100.00} & \textbf{100.00} & 3.48 \\
				FNN(20) & \textbf{100.00} & \textbf{100.00} & \textbf{100.00} & 2.47 \\
				\bottomrule
			\end{tabular}
		\end{small}
	\end{center}
	\vskip -0.1in
\end{table}

\subsection{Real Data Experiments (EEG Data)}
EEG is a non-invasive method for measuring electrical activity of the brain via electrodes placed on the scalp. One of the major challenges of analyzing EEG data is the presence of artifacts, such as eye movements and muscle activity. Further, EEG signals are typically noisy and complex, making it difficult to identify meaningful patterns and signals. Moreover, when brain-computer interfaces based on EEG signals are developed, that use self-paced actions rather than externally triggered signals, curve registration becomes infeasible, thus adding to the challenge. In the following, we show that the proposed methodology is able to handle these challenges.

\subsubsection*{Setup}
The BCI Competition IV Dataset 2A \citep{tangermann2012} is a common benchmark for evaluating the performance of a new method for the analysis of EEG data. The dataset consists of the recorded brain activity of 9 participants. More specifically, for each participant two sessions of approximately 45 minutes each were recorded on different days. As usual in the literature, we used the first recording per participant as training and the second as validation data.

According to the documentation, participants were asked to imagine movements of their left hand (class 1), right hand (class 2), both feet (class 3) and tongue (class 4). During each session, every imaginary movement was repeated 72 times, yielding a total of 288 trials. Each trial took approximately 8 seconds. At the beginning of each trial ($t=0\,s$), a short acoustic signal and a fixation cross on a black screen appeared. Two seconds later ($t=2\,s$), a visual cue appeared to indicate the movement, which should be imagined. The imaginary movement can be assumed to start approximately half a second after the cue ($t=2.5\,s$) and end when the fixation cross disappeared ($t=6\,s$). Each trial was followed by a short break to separate it from subsequent trials. The participants' brain activity was measured through a 22-channel EEG with 3 additional EOG (\textit{Electrooculography}) channels at a sampling rate of 250\,Hz.

For this dataset, the \textit{classic approach} to benchmark a new method is to cut windows from each trial, e.\,g., between 2.5\,s and 4.5\,s after trial onset, which is feasible since the trial and cue onsets are known. However, if we move beyond externally triggered actions, we need another approach. This is particularly important in the case of brain-computer interfaces where devices should be controlled continuously. In this case, a common approach is to use \textit{sliding windows}, i.\,e., to use overlapping windows of a fixed length with a fixed step size. 

We tested the proposed functional neural network, as described in Section \ref{sec:architectures} with the same specifications as in the simulation study, and compared it to the EEGNet with its default choice of hyperparameters as suggested by Lawhern et~al. \yrcite{lawhern2018}. However, to account for the different degrees of complexity of the EEG data, we chose to use different numbers of filters in the convolutional layers. For the classic approach with 4 classes (corresponding to the 4 different imaginary movements), we tested models with 40 + 20, 5 + 10, 3 + 12 and 20 filters, respectively, and denote these models by FNN(40, 20), FNN(5, 10), FNN(3, 12) and FNN(20). Further, we used an F-MLP with a single hidden layer of 20 functional neurons. The numbers of parameters are displayed in Table \ref{tab:real_data_results_classic}. For the sliding window approach with 7 classes, we similarly used the models FNN(40, 20), FNN(5, 10), FNN(20) and FNN(40), with their according number of parameters displayed in Table \ref{tab:real_data_results_sliding}.

For the classic approach, we used windows between the cue onset ($t=2\,s$) and the disappearance of the fixation cross ($t=6\,s$). We trained the proposed FNN and the EEGNet with 2,250 batches of size 32 to distinguish between the four classes (left hand, right hand, feet, tongue). In total 72,000 samples were used, which means that each of the 288 training windows was used 250 times. 

For the sliding windows approach, we used windows of 1\,s and a step size of 0.004\,s which led to approximately 675,000 sliding windows. These windows might coincide with a break between trials (class 1), the time between trial and cue onset (class 3), the time between cue onset and imagined movement (class 2) or one of the four imagined movements (classes 4 - 7). The windows at the transition between two classes were labeled with the most frequent class. This problem is substantially more complex than the classic approach and we have more varied data. We trained both models with 16,000 batches of size 32, thus 512,000 windows, which is close to the total number of sliding windows. We trained each model 10 times for each of the 9 recordings.

\subsubsection*{Results}
The results for both approaches are displayed in Tables \ref{tab:real_data_results_classic} and \ref{tab:real_data_results_sliding}. As before, the accuracy represents the ratio of correctly classified windows to all windows. To account for the class imbalance, the mean recall and precision over all categories were added: the recall of a binary classifier is defined as the ratio of correctly classified positive to all true positive samples, whereas the precision of a binary classifier is defined as the ratio of correctly classified positive to all positively classified samples. These quantities for binary classifiers were extended to the multiclass problem by first calculating the respective quantity per class and then averaging the calculated quantities over all classes.

For the classic approach, the benchmark model outperformed the FNNs in terms of all metrics. It is however notably, that the smaller FNNs with 2,344 and 1,564 parameters, compared to the EEGNet's 6,980 parameters, still achieve competitive results. The respective confusion matrices are displayed in Figure \ref{cm:classic} of Appendix \ref{app:real_data}.

For the sliding window approach, the FNN(40) achieves the best performance, and specifically outperforms the benchmark model with a 4.48\% higher accuracy. It is also worth to mention that the smallest model, the FNN(5, 10) with approximately 10\% less parameters still achieves better results in terms of accuracy and recall compared to the benchmark model. Note that classes 4 - 7 are generally more difficult to detect. Yet, it can be seen from the confusion matrices in Figure \ref{cm:sliding_win} of Appendix \ref{app:real_data}, that the classifications of the FNNs with a single convolutional layer are particularly better for those classes, which is also reflected in the recall and precision of the classifiers.

Note that the functional layers were specifically designed for the sliding window approach, where curve registration is not feasible. In the classic case, it might be sufficient to use functional dense layers rather than convolutions. This is supported by the competitive performance of the F-MLP(20), which consists of a single hidden layer with 20 functional neurons as defined in \eqref{eq:neurons}.

Further, the results seem to be rather stable. From Table \ref{tab:accuracy_per_pax} of Appendix \ref{app:real_data} it can be seen that the differences between the results are generally of a similar order as the variability between different training runs. In particular, the differences between the EEGNet and the FNNs with a single convolutional layer are larger than the variability of the results, indicating that FNNs are generally preferably compared to the benchmark model. 

The computational complexity of all models is similar, as can be seen from Table \ref{tab:computational_complexity} of Appendix \ref{app:real_data}. For example, training the EEGNet under the classic approach with 2,500 batches took 25.8s ($\pm 0.9$) compared to 27.9s ($\pm 0.3$) for the FNN(5, 10). The time complexity is based on an implementation in TensorFlow and training on an NVIDIA GeForce GTX 1650 GPU and an Intel(R) Core(TM) i5-9300H @ 2.40GHz CPU.

Both the FNNs and the (default) EEGNet are relatively simple models. It can be expected that the accuracies improve for both types of models if the hyperparameters are tuned carefully. Further improvements might be possible by changing the FNN's architecture or simply using more layers.
\begin{table}
	\caption{Comparison of the models' qualities and number of parameters under the classic approach.}
	\label{tab:real_data_results_classic}
	\vskip 0.1in
	\begin{center}
		\begin{small}
			\begin{tabular}{lcccc}
				\toprule
				Model & Accuracy & Recall & Precision & Parameters \\
				\midrule
				EEGNet & \textbf{72.13} & \textbf{72.14} & \textbf{72.27} & 6,980 \\
				FNN(40, 20) & 69.08 & 69.06 & 69.04 & 19,464 \\
				FNN(5, 10) & 65.51 & 65.49 & 65.58 & 2,344 \\
				FNN(3, 12) & 64.32 & 64.32 & 64.35 & 1,564 \\
				FNN(20) & 67.23 & 67.25 & 67.21 & 7,924 \\
				F-MLP(20) & 66.31 & 66.31 & 66.33 & 7,924 \\
				\bottomrule
			\end{tabular}
		\end{small}
	\end{center}
	\vskip -0.1in
\end{table}
\begin{table}
	\caption{Comparison of the models' qualities and number of parameters under the sliding windows approach.}
	\label{tab:real_data_results_sliding}
	\vskip 0.1in
	\begin{center}
		\begin{small}
			\begin{tabular}{lcccc}
				\toprule
				Model & Accuracy & Recall & Precision & Parameters \\
				\midrule
				EEGNet & 51.81 & 40.21 & 44.75 & 2,783 \\
				FNN(40, 20) & 51.70 & 40.64 & 43.39 & 19,767 \\
				FNN(5, 10) & 52.12 & 40.54 & 44.32 & 2,497 \\
				FNN(20) & 55.43 & 44.98 & 48.31 & 8,227 \\
				FNN(40) & \textbf{56.29} & \textbf{45.95} & \textbf{49.46} & 16,447 \\
				Func2Func(2) & 51.52 & 36.97 & 48.14 & 8,227 \\
				Func2Func(3) & 47.33 & 30.48 & 39.70 & 8,887 \\
				\bottomrule
			\end{tabular}
		\end{small}
	\end{center}
	\vskip -0.1in
\end{table}

\subsubsection*{Fully Functional Predictions}

With the proposed methodology it is not only possible to predict scalar-valued labels and use the model for classification, but it is also possible to predict functional labels. With the sliding windows as before, we can try to predict the class label for each time point rather than one label for the whole window. This is particularly useful at the transition from one state to another because these transitions cannot be represented by a simple classification. Note that such per-time predictions are also possible with specialized neural networks with conventional layers, but these generally do not preserve the smoothness of the data.

\begin{figure}[ht]
	\vskip 0.1in
	\begin{center}
		\centerline{\includegraphics[width=\columnwidth]{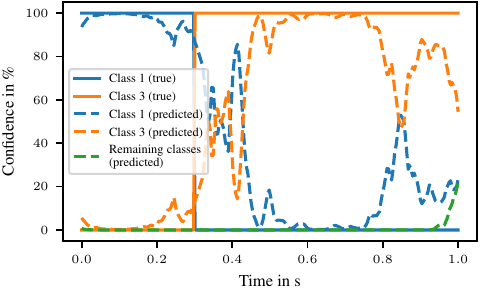}}
		\caption{True and predicted class labels for a window of one second at the transition from class 1 (inter-trial break) to class 3 (time after trial onset).}
		\label{fig:fully_func_pred_1_3}
	\end{center}
	\vskip -0.3in
\end{figure}

\begin{figure}[ht]
	\vskip 0.1in
	\begin{center}
		\centerline{\includegraphics[width=\columnwidth]{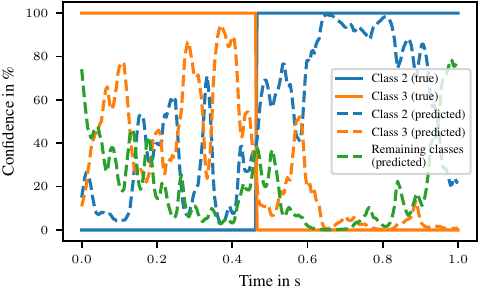}}
		\caption{True and predicted class labels for a window of one second at the transition from class 3 (time after trial onset) to class 2 (time after cue onset).}
		\label{fig:fully_func_pred_3_2}
	\end{center}
	\vskip -0.3in
\end{figure}

We trained two FNNs with two and three functional convolutional layers, referred to as Func2Func(2) and Func2Func(3) respectively, to predict labels for each time instant of the sliding windows. Model Func2Func(3) is depicted on the right of Figure \ref{fig:architecture}. Aggregated results are displayed in Table \ref{tab:real_data_results_sliding} (and more detailed in Figure \ref{cm:func_2_func} of Appendix \ref{app:real_data}). Here, a single classification is obtained from a window by selecting the class with the highest summed confidence. The aggregated results are slightly worse compared to the scalar classifiers above. In Figures \ref{fig:fully_func_pred_1_3} and \ref{fig:fully_func_pred_3_2} are the true and predicted labels of two windows at the transition from an inter-trial break (class 1) to the interval after a trial onset (class 3) and from the interval after a trial onset (class 3) to the interval after a cue onset (class 2). It can be seen from both figures that the predictions do not match the true labels perfectly and that the confidences at the border region are generally lower, but overall the predictions match the true labels.

Thus, even though the aggregated results are slightly worse, fully functional predictions give more detailed insights into the windows and avoid the aggregation of labels per time step to a single label, which is particularly useful at the transition between two classes.

\section{Conclusions}  \label{sec:con}

In this work, we presented a new framework to analyze functional data with scalar- or functional-valued targets. We combined advantages of convolutional neural networks with those of functional data analysis. More specifically, we proposed a neural network that can be considered as shift invariant while taking the intrinsic functional structure of the data into account. 

We showed that even shallow models with only two convolutional and one dense layer are more powerful than the functional k-nearest neighbors algorithm for the simulated data. Further, the results of our case study suggest that FNNs with a similar amount of trainable weights outperform EEGNet, the de facto standard model for the classification of EEG data.

The results of this paper suggest that functional neural networks are a relevant area for future research. First, it could be tested if FNNs work well with other types of time series and functional data, such as stock prices or temperature curves. It might be interesting to investigate if the methodology can be expanded to other types of data like images (considered as functions of the two variables height and width) or videos (considered as functions of the three variables time, height and width). Further, the choice of base is crucial for the performance of the network. Prior simulation studies showed that the Fourier base and Legendre polynomials lead to good results, but other bases might further improve the predictions. In FDA it is common to find roughness penalties as regularizers. Although a preliminary simulation study suggested that a base representation of the weight functions leads to better results, it would be interesting to study if the weight functions in the neural network can be learned directly while their smoothness would be ensured via corresponding roughness penalties. Finally, the proposed framework could be extended to other types of neural networks, such as recurrent neural networks or transformers.

\section*{Acknowledgment} 

This work is supported by the Ministry of Economics, Innovation, Digitization and Energy of the State of North Rhine-Westphalia and the European Union, grant IT-2-2-023 (VAFES). We would like to thank anonymous reviewers for their valuable comments and constructive feedback.

\bibliography{bibliography}
\bibliographystyle{icml2023}

\newpage
\appendix
\onecolumn

\section{Local Linear Estimation - Calculations}
\label{app:loc_lin}

Explicit formulas can be derived via straightforward calculations from the definition of the local polynomial estimator. In the following, we derive an explicit formula of the local linear estimator, where $p=1$. In this case, the definition in \eqref{eq:loc_lin} simplifies to $(\hat{f}(x), \widehat{f'}(x)) = \argmin_{\beta_0, \beta_1} F(\beta_0, \beta_1)$, where

$$
	F(\beta_0, \beta_1) = \sum_{t=1}^{T} \bigg(X_t - \beta_0 - \beta_1 \Big(\frac{t}{T}-x\Big)\bigg)^2 K_h\big(\frac{t}{T}-x\big).
$$

Now, define
$$ R_k(x) = \sum_{t=1}^{T} X_t \Big(\frac{t}{T}-x\Big)^k K_h\big(\frac{t}{T}-x\big)\quad \text{and} \quad S_k(x) = \sum_{t=1}^{T} \Big(\frac{t}{T}-x\Big)^k K_h\big(\frac{t}{T}-x\big), $$
for $k = 0, 1, 2$. Then, we can calculate the partial derivatives of $F$ as 
$$ \frac{\partial F}{\partial \beta_j} = -2 \big(R_j(x) - \beta_0 S_j(x) - \beta_1 S_{j+1}(x)\big) \quad \text{and} \quad \frac{\partial^2 F}{\partial \beta_j \partial \beta_k} = 2 S_{j+k}(x) $$
for $j, k\in \{0, 1\}$. Note that $(hT)^{-1} S_k(x) \xrightarrow{hT\to\infty} \int_{-1}^1 x^k K(x) \diff x$, thus $S_0(x), S_2(x) > 0$ and by symmetry of $K$, $(hT)^{-1} S_1(x)\approx 0$, for $hT$ sufficiently large. By assumption $\int_{-1}^1K(x)\diff x > \int_{-1}^1x^2 K(x)\diff x$, thus it follows that $S_0(x) > S_2(x)$ and therefore that the Hessian matrix
$$ \mathbf{H}_F = 2 \begin{bmatrix}
	S_0(x) & S_1(x) \\ S_1(x) & S_2(x)
\end{bmatrix} $$
is positive definite (and in particular invertible) for sufficiently large values of $hT$. 

Setting the gradient $\big(\frac{\partial F}{\partial \beta_0}, \frac{\partial F}{\partial \beta_1}\big)$ equal to zero, we obtain the equation

$$ \begin{bmatrix} S_0(x) & S_1(x) \\ S_1(x) & S_2(x) \end{bmatrix} 
	\cdot \begin{bmatrix} \beta_0 \\ \beta_1 \end{bmatrix} 
	= \begin{bmatrix} R_0(x) \\ R_1(x) \end{bmatrix}. $$ 

Solving this equation for $(\beta_0, \beta_1)$, we finally obtain the local linear estimator

$$\beta_k = \frac{(-1)^kS_{2-k}(x)R_0(x) - (-1)^k S_{1-k}(x)R_1(x)}{S_2(x)S_0(x) - S_1^2(x)},$$
for $k=0, 1$. This formula can be further simplified and efficiently implemented.

\section{Simulated Data - Samples}\label{app:sim_data}

\begin{figure}[H]
	\vskip -0.1in
	\begin{center}
		\includegraphics[width=0.45\textwidth]{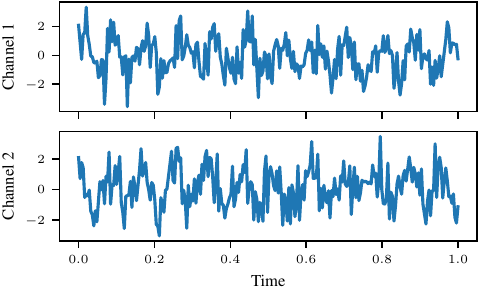} \hspace{0.5cm}
		\includegraphics[width=0.45\textwidth]{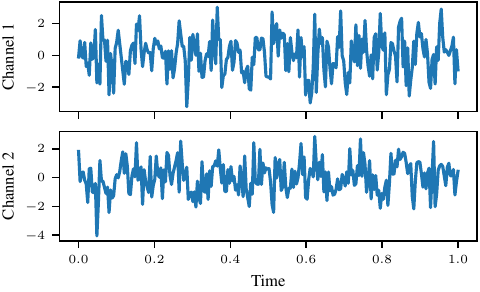}\\ 
		\includegraphics[width=0.45\textwidth]{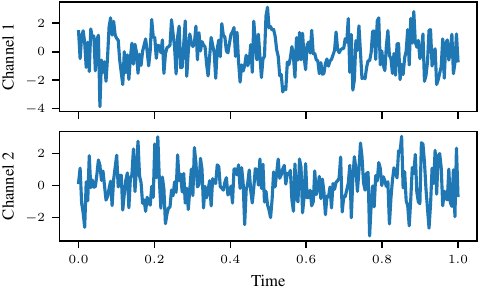}
		\vskip -0.1in
		\caption{Samples from the simulated dataset I. Top left: Example of class 1 ($\alpha$ frequencies only). Top right: Example of class 2 ($\beta$ frequencies added in both channels, stronger in Channel 1). Bottom: Example of class 3 ($\beta$ frequencies added in both channels, stronger in Channel 2).}
		\label{fig:dataset1}
	\end{center}
\end{figure}

\begin{figure}[H]
	\vskip -0.2in
	\begin{center}
		\includegraphics[width=0.45\textwidth]{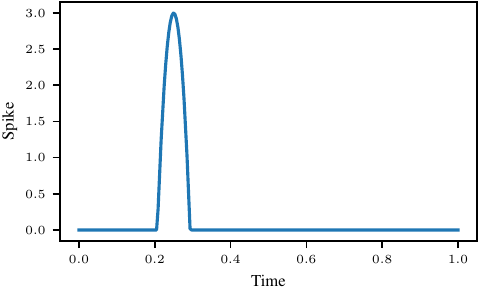} \hspace{0.5cm}
		\includegraphics[width=0.45\textwidth]{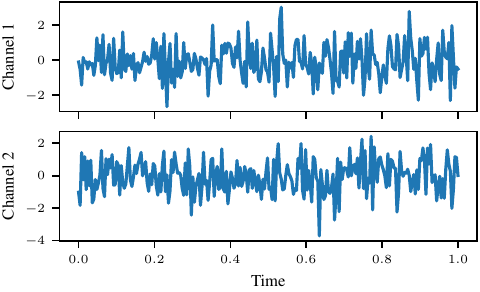} \\
		\includegraphics[width=0.45\textwidth]{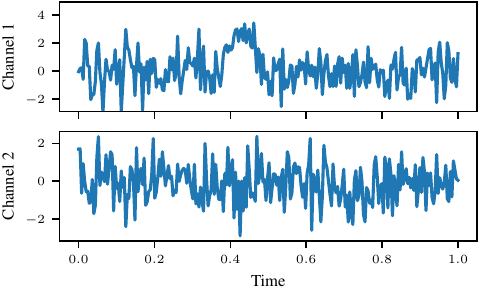} \hspace{0.5cm}
		\includegraphics[width=0.45\textwidth]{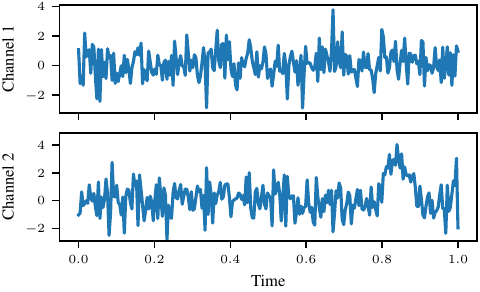}
		\vskip -0.1in
		\caption{Samples from the simulated dataset II. Top left: ``Spike'' at $t=0.25$ with width $w=0.05$. Top right: Example of class 1 (white noise in both channels). Bottom left: Example of class 2 (``spike'' added to white noise in Channel 1). Bottom right: Example of class 3 (``spike'' added to white noise in Channel 2).}
		\label{fig:dataset2}
	\end{center}
\end{figure}

\section{Simulated Data - Results}\label{app:sim_data_res}

\begin{figure}[ht]
	\vskip 0.2in
	\begin{center}
		\includegraphics[width=0.48\textwidth]{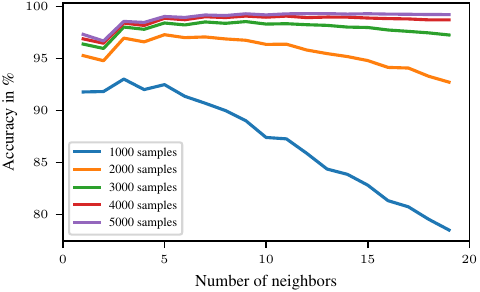} \hspace{0.02\textwidth} 
		\includegraphics[width=0.48\textwidth]{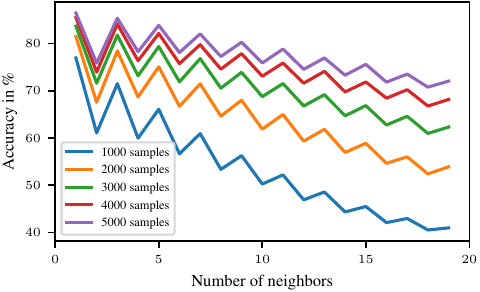}
		\caption{Mean accuracy in percent (y-axis) of the $k$-nearest neighbors classifiers for a varying number of neighbors (x-axis). Left: Dataset I. Right: Dataset II.}
		\label{fig:simulation_results}
	\end{center}
	\vskip -0.2in
\end{figure}

\section{Real Data Experiments (EEG Data) - Results}\label{app:real_data}

\begin{figure}[H]
	\vskip -0.2in
	\begin{center}
		\centerline{{\small\begin{tikzpicture}[scale=0.75]	
	\foreach \y [count=\n] in {
		{17.5, 2.9, 3.5, 1.4},
		{2.0, 18.9, 2.5, 1.4},
		{2.9, 2.3, 17.1, 2.5},
		{1.3, 2.0, 3.0, 18.7},
	} {
			\ifnum\n<5
			\node[minimum size=8mm] at (\n, 0) {\n};
			\fi
			\foreach \x [count=\m] in \y {
					\node[fill=black!\x!white, minimum size=8mm, text=black] at (\m,-\n) {\x};
				}
	()	}
	
	\foreach \a [count=\i] in {1,2,3,4} {
			\node[minimum size=8mm] at (0,-\i) {\a};
		}
	
	\begin{scope}[xshift=5.5cm]	
		\foreach \y [count=\n] in {
			{17.2, 3.1, 3.2, 1.5},
			{2.4, 19.3, 2.3, 1.2},
			{2.7, 3.2, 15.0, 4.0},
			{2.0, 2.4, 3.0, 17.6},
		} {
			\ifnum\n<5
			\node[minimum size=8mm] at (\n, 0) {\n};
			\fi
			\foreach \x [count=\m] in \y {
				\node[fill=black!\x!white, minimum size=8mm, text=black] at (\m,-\n) {\x};
			}
			()	}
		
		\foreach \a [count=\i] in {1,2,3,4} {
			\node[minimum size=8mm] at (0,-\i) {\a};
		}
	\end{scope}
	
	\begin{scope}[xshift=11.0cm]	
		\foreach \y [count=\n] in {
			{15.9, 2.9, 4.2, 2.0},
			{2.3, 18.7, 2.8, 1.3},
			{2.7, 3.2, 14.6, 4.3},
			{2.5, 2.0, 4.2, 16.3},
		} {
			\ifnum\n<5
			\node[minimum size=8mm] at (\n, 0) {\n};
			\fi
			\foreach \x [count=\m] in \y {
				\node[fill=black!\x!white, minimum size=8mm, text=black] at (\m,-\n) {\x};
			}
			()	}
	
		\foreach \a [count=\i] in {1,2,3,4} {
			\node[minimum size=8mm] at (0,-\i) {\a};
		}
	\end{scope}

	\begin{scope}[yshift=-5.5cm]	 
		\foreach \y [count=\n] in {
			{15.6, 3.4, 4.0, 2.0},
			{2.5, 18.1, 2.7, 1.8},
			{2.9, 3.1, 14.5, 4.6},
			{2.2, 2.3, 4.2, 16.2},
		} {
		\ifnum\n<5
		\node[minimum size=8mm] at (\n, 0) {\n};
		\fi
		\foreach \x [count=\m] in \y {
			\node[fill=black!\x!white, minimum size=8mm, text=black] at (\m ,-\n) {\x};
		}
		()	}
	
		\foreach \a [count=\i] in {1,2,3,4} {
			\node[minimum size=8mm] at (0,-\i) {\a};
		}
	\end{scope}

	\begin{scope}[shift={(5.5cm,-5.5cm)}]	 
		\foreach \y [count=\n] in {
			{16.4, 3.1, 3.9, 1.8},
			{2.1, 18.9, 2.5, 1.4},
			{2.7, 3.1, 14.8, 4.4},
			{2.1, 2.2, 3.5, 17.2},
		} {
			\ifnum\n<5
			\node[minimum size=8mm] at (\n, 0) {\n};
			\fi
			\foreach \x [count=\m] in \y {
				\node[fill=black!\x!white, minimum size=8mm, text=black] at (\m ,-\n) {\x};
			}
			()	}
		
		\foreach \a [count=\i] in {1,2,3,4} {
			\node[minimum size=8mm] at (0,-\i) {\a};
		}
	\end{scope}

	\begin{scope}[shift={(11.0cm,-5.5cm)}]	 
		\foreach \y [count=\n] in {
			{16.0, 2.9, 4.2, 1.9},
			{3.2, 17.7, 2.4, 1.6},
			{3.2, 2.9, 15.2, 3.6},
			{2.1, 2.1, 3.5, 17.4},
		} {
			\ifnum\n<5
			\node[minimum size=8mm] at (\n, 0) {\n};
			\fi
			\foreach \x [count=\m] in \y {
				\node[fill=black!\x!white, minimum size=8mm, text=black] at (\m ,-\n) {\x};
			}
			()	}
		
		\foreach \a [count=\i] in {1,2,3,4} {
			\node[minimum size=8mm] at (0,-\i) {\a};
		}
	\end{scope}
	
\end{tikzpicture}
}}
		\caption{Confusion matrices for different models under the classic approach. Top left: EEGNet. Top center: FNN(40, 20). Top right: FNN(5, 10). Bottom left: FNN(3, 12). Bottom center: FNN(20). Bottom right: F-MLP(20).}
		\label{cm:classic}
	\end{center}
	\vskip -0.2in
\end{figure}

\begin{figure}[H]
	\vskip -0.2in
	\begin{center}
		\centerline{{\footnotesize\begin{tikzpicture}[scale=0.7] 		
	\foreach \y [count=\n] in {
		{27.1, 0.7, 4.2, 0.8, 1.0, 0.9, 0.9},
		{1.6, 5.8, 2.2, 0.4, 0.4, 0.4, 0.3},
		{6.6, 1.3, 10.7, 0.6, 0.7, 0.7, 1.0},
		{2.5, 0.5, 1.0, 2.2, 1.0, 0.5, 0.3},
		{2.6, 0.3, 1.4, 0.6, 2.3, 0.4, 0.4},
		{2.1, 0.4, 1.7, 0.4, 0.4, 1.9, 1.0},
		{2.6, 0.4, 1.7, 0.4, 0.3, 0.9, 1.9},
	} {
		\ifnum\n<8
		\node[minimum size=8mm] at (\n, 0) {\n};
		\fi
		\foreach \x [count=\m] in \y {
			\node[fill=black!\x!white, minimum size=8mm, text=black] at (\m,-\n) {\x};
		}
		()	}
	
	\foreach \a [count=\i] in {1,2,3,4,5,6,7} {
		\node[minimum size=8mm] at (0,-\i) {\a};
	}

\begin{scope}[xshift=8.5cm] 
	\foreach \y [count=\n] in {
		{26.2, 0.6, 3.6, 1.4, 1.4, 1.2, 1.2},
		{1.4, 6.6, 1.6, 0.3, 0.3, 0.4, 0.3},
		{6.1, 1.0, 10.9, 0.9, 0.9, 0.8, 1.0},
		{2.4, 0.3, 1.1, 2.1, 1.1, 0.7, 0.4},
		{2.5, 0.3, 1.0, 0.9, 2.1, 0.7, 0.4},
		{2.3, 0.3, 1.4, 0.6, 0.7, 1.8, 0.9},
		{2.2, 0.3, 1.3, 0.5, 0.5, 1.0, 1.9},
	} {
		\ifnum\n<8
		\node[minimum size=8mm] at (\n, 0) {\n};
		\fi
		\foreach \x [count=\m] in \y {
			\node[fill=black!\x!white, minimum size=8mm, text=black] at (\m,-\n) {\x};
		}
		()	}
	
	\foreach \a [count=\i] in {1,2,3,4,5,6,7} {
		\node[minimum size=8mm] at (0,-\i) {\a};
	}
\end{scope}

\begin{scope}[xshift=17.0cm] 
	\foreach \y [count=\n] in {
		{27.2, 0.7, 3.8, 0.9, 1.3, 0.8, 1.1},
		{1.4, 6.5, 1.8, 0.3, 0.3, 0.2, 0.2},
		{6.7, 1.4, 10.5, 0.7, 0.8, 0.5, 1.0},
		{2.6, 0.5, 1.1, 2.0, 1.0, 0.5, 0.3},
		{2.7, 0.4, 1.2, 0.6, 2.2, 0.5, 0.4},
		{2.5, 0.4, 1.4, 0.5, 0.6, 1.4, 1.1},
		{2.7, 0.3, 1.3, 0.3, 0.4, 0.8, 2.1},
	} {
		\ifnum\n<8
		\node[minimum size=8mm] at (\n, 0) {\n};
		\fi
		\foreach \x [count=\m] in \y {
			\node[fill=black!\x!white, minimum size=8mm, text=black] at (\m,-\n) {\x};
		}
		()	}
	
	\foreach \a [count=\i] in {1,2,3,4,5,6,7} {
		\node[minimum size=8mm] at (0,-\i) {\a};
	}
\end{scope}

\begin{scope}[shift={(4.25cm,-8.5cm)}] 
	\foreach \y [count=\n] in {
		{27.5, 0.6, 3.7, 1.0, 1.4, 1.1, 1.0},
		{1.3, 6.6, 1.6, 0.3, 0.3, 0.4, 0.3},
		{5.7, 1.1, 11.6, 0.6, 0.8, 0.7, 1.0},
		{2.2, 0.4, 0.9, 2.4, 1.0, 0.6, 0.3},
		{2.1, 0.3, 1.0, 0.7, 2.7, 0.6, 0.3},
		{2.2, 0.4, 1.4, 0.4, 0.5, 2.2, 0.9},
		{2.1, 0.3, 1.3, 0.4, 0.4, 1.0, 2.4},
	} {
		\ifnum\n<8
		\node[minimum size=8mm] at (\n, 0) {\n};
		\fi
		\foreach \x [count=\m] in \y {
			\node[fill=black!\x!white, minimum size=8mm, text=black] at (\m,-\n) {\x};
		}
		()	}
	
	\foreach \a [count=\i] in {1,2,3,4,5,6,7} {
		\node[minimum size=8mm] at (0,-\i) {\a};
	}
\end{scope}

\begin{scope}[shift={(12.75cm,-8.5cm)}] 
	\foreach \y [count=\n] in {
		{27.7, 0.6, 3.4, 1.1, 1.2, 1.0, 1.1},
		{1.3, 6.8, 1.5, 0.2, 0.3, 0.3, 0.3},
		{5.8, 1.1, 11.8, 0.7, 0.7, 0.5, 0.9},
		{2.2, 0.4, 0.8, 2.6, 0.9, 0.6, 0.3},
		{2.3, 0.4, 1.0, 0.7, 2.7, 0.6, 0.3},
		{2.1, 0.3, 1.2, 0.4, 0.6, 2.3, 1.0},
		{2.3, 0.3, 1.2, 0.4, 0.4, 0.9, 2.5},
	} {
		\ifnum\n<8
		\node[minimum size=8mm] at (\n, 0) {\n};
		\fi
		\foreach \x [count=\m] in \y {
			\node[fill=black!\x!white, minimum size=8mm, text=black] at (\m,-\n) {\x};
		}
		()	}
	
	\foreach \a [count=\i] in {1,2,3,4,5,6,7} {
		\node[minimum size=8mm] at (0,-\i) {\a};
	}
\end{scope}
\end{tikzpicture}
}}
		\caption{Confusion matrices for different models under the sliding window approach. Top left: EEGNet. Top center: FNN(40, 20). Top right: FNN(5, 10). Bottom left: FNN(20). Bottom right FNN(40).}
		\label{cm:sliding_win}
	\end{center}
	\vskip -0.2in
\end{figure}

\begin{figure}[H]
	\vskip -0.2in
	\begin{center}
		\centerline{{\footnotesize\begin{tikzpicture}[scale=0.7] 		
	\foreach \y [count=\n] in {
		{28.7, 0.5, 4.6, 0.5, 0.7, 0.2, 0.5}, 
		{1.9, 5.8, 2.4, 0.2, 0.3, 0.0, 0.2},
		{7.8, 1.3, 11.2, 0.2, 0.3, 0.1, 0.4},
		{2.9, 0.5, 1.9, 1.7, 0.5, 0.2, 0.2},
		{3.0, 0.4, 2.3, 0.4, 1.8, 0.1, 0.1},
		{3.2, 0.5, 2.4, 0.2, 0.2, 0.6, 0.8},
		{3.2, 0.5, 2.2, 0.1, 0.1, 0.3, 1.7},
	} {
		\ifnum\n<8
		\node[minimum size=8mm] at (\n, 0) {\n};
		\fi
		\foreach \x [count=\m] in \y {
			\node[fill=black!\x!white, minimum size=8mm, text=black] at (\m,-\n) {\x};
		}
		()	}
	
	\foreach \a [count=\i] in {1,2,3,4,5,6,7} {
		\node[minimum size=8mm] at (0,-\i) {\a};
	}

\begin{scope}[xshift=8.5cm] 
	\foreach \y [count=\n] in {
		{29.5, 0.5, 4.5, 0.3, 0.5, 0.2, 0.4},
		{2.4, 5.1, 2.7, 0.1, 0.2, 0.1, 0.2},
		{9.3, 1.6, 9.6, 0.2, 0.3, 0.1, 0.4},
		{3.4, 0.6, 2.5, 0.7, 0.5, 0.2, 0.2},
		{3.4, 0.4, 2.4, 0.3, 1.1, 0.2, 0.1},
		{3.5, 0.6, 2.3, 0.2, 0.3, 0.3, 0.5},
		{3.6, 0.4, 2.6, 0.1, 0.2, 0.2, 0.9},
	} {
		\ifnum\n<8
		\node[minimum size=8mm] at (\n, 0) {\n};
		\fi
		\foreach \x [count=\m] in \y {
			\node[fill=black!\x!white, minimum size=8mm, text=black] at (\m,-\n) {\x};
		}
		()	}
	
	\foreach \a [count=\i] in {1,2,3,4,5,6,7} {
		\node[minimum size=8mm] at (0,-\i) {\a};
	}
\end{scope}

\end{tikzpicture}
}}
		\caption{Confusion matrices for different fully functional models under the sliding window approach. The functional predictions were aggregated to a single scalar prediction by using the class with the highest average confidence. Left: Func2Func(2). Right: Func2Func(3).}
		\label{cm:func_2_func}
	\end{center}
	\vskip -0.2in
\end{figure}

\begin{table}[H]
	\caption{Comparison of the models' variability across different training runs and recordings. Every model was trained 10 times and the average accuracy is displayed in the table (standard deviation in parentheses).}
	\label{tab:accuracy_per_pax}
	\vskip 0.1in
	\begin{center}
		\begin{small}
			\begin{tabular}{lccccc}
				\toprule
				Participant & EEGNet & FNN(40, 20) & FNN(5, 10) & FNN(20) & FNN(40) \\
				\midrule				
				1 & 49.16 ($\pm$ 1.50) & 49.16 ($\pm$ 1.39) & 49.41 ($\pm$ 1.99) & 52.19 ($\pm$ 1.14) & 54.73 ($\pm$ 1.70) \\ 
				2 & 38.24 ($\pm$ 1.98) & 40.14 ($\pm$ 1.16) & 40.14 ($\pm$ 0.91) & 42.73 ($\pm$ 1.26) & 43.18 ($\pm$ 0.99) \\
				3 & 55.25 ($\pm$ 1.13) & 56.56 ($\pm$ 0.91) & 54.38 ($\pm$ 1.53) & 59.00 ($\pm$ 1.96) & 59.82 ($\pm$ 0.85) \\
				4 & 52.52 ($\pm$ 1.68) & 55.02 ($\pm$ 0.99) & 55.21 ($\pm$ 1.11) & 58.73 ($\pm$ 1.02) & 60.33 ($\pm$ 1.12) \\
				5 & 49.49 ($\pm$ 3.74) & 49.55 ($\pm$ 1.28) & 51.68 ($\pm$ 1.21) & 54.55 ($\pm$ 0.74) & 53.46 ($\pm$ 1.87) \\
				6 & 48.42 ($\pm$ 1.96) & 47.83 ($\pm$ 1.23) & 50.21 ($\pm$ 1.42) & 52.52 ($\pm$ 1.08) & 52.50 ($\pm$ 1.33) \\
				7 & 61.46 ($\pm$ 0.92) & 61.39 ($\pm$ 1.21) & 58.65 ($\pm$ 1.86) & 64.20 ($\pm$ 1.45) & 65.18 ($\pm$ 2.82) \\
				8 & 55.16 ($\pm$ 1.34) & 54.04 ($\pm$ 2.04) & 52.15 ($\pm$ 2.75) & 56.62 ($\pm$ 0.55) & 59.12 ($\pm$ 1.98) \\
				9 & 50.82 ($\pm$ 1.19) & 44.96 ($\pm$ 1.77) & 50.66 ($\pm$ 2.34) & 50.62 ($\pm$ 0.83) & 51.07 ($\pm$ 1.54) \\
				\bottomrule
			\end{tabular}
		\end{small}
	\end{center}
	\vskip -0.1in
\end{table}

\begin{table}[H]
	\caption{Comparison of the models' computational complexity. Time for 2500 and 4000 steps of training for the classic and sliding window approach, respectively. Mean time calculated over 10 repetitions (standard deviation in parentheses).}
	\label{tab:computational_complexity}
	\vskip 0.1in
	\begin{center}
		\begin{small}
			\begin{tabular}{lcc}
				\toprule
				Model & 'classic' approach (2500 steps) & 'sliding window' approach (4000 steps) \\
				\midrule
				EEGNet & 25.8s ($\pm$ 0.9) & 40.2s ($\pm$ 0.6) \\
				FNN(5, 10) & 27.9s ($\pm$ 0.3) &  47.3s ($\pm$ 2.6) \\
				FNN(3, 12) & 28.0s ($\pm$ 0.6) & 48.24 ($\pm$ 1.5) \\
				\bottomrule
			\end{tabular}
		\end{small}
	\end{center}
	\vskip -0.1in
\end{table}

\end{document}